\documentclass{article}
\usepackage[utf8]{inputenc}
\usepackage[T1]{fontenc}
\usepackage[expansion=false]{microtype}
\usepackage{graphicx}
\usepackage{booktabs}
\usepackage{multirow}
\usepackage{array}
\usepackage{tabularx}
\usepackage{float}
\usepackage{stfloats}
\usepackage{amsmath}
\usepackage{amssymb}
\usepackage[table]{xcolor}
\usepackage{hyperref}
\usepackage{xurl}
\hypersetup{hypertexnames=false}

\usepackage[accepted]{mlsys2025}
\makeatletter
\renewcommand{\Notice@String}{Preprint.}
\makeatother
\providecommand{\tablefont}{}

\mlsystitlerunning{SpecZip: Efficient Speculative Drafter Training on Compressed Hidden-State Caches}

\hypersetup{pdftitle={SpecZip: Efficient Speculative Drafter Training on Compressed Hidden-State Caches},pdfauthor={Minsub Kim, Jaeeun Kil, Sungjae Lee, Jiwon Ryu, Jeongtae Lee, Hanbae Seo, Hoseung Kim, Sunghwan Shim, Jaegwang Lee, Hyunjoon Jeong, Junhee Yoo, Sunghyeon Woo}}
\begin{document}

\twocolumn[
\mlsystitle{SpecZip: Efficient Speculative Drafter Training \\ on Compressed Hidden-State Caches}

\mlsyssetsymbol{equal}{*}
\mlsyssetsymbol{corr}{\textdagger}
\begin{mlsysauthorlist}
\mlsysauthor{Minsub Kim}{equal,nc}
\mlsysauthor{Jaeeun Kil}{equal,nc}
\mlsysauthor{Sungjae Lee}{nc}
\mlsysauthor{Jiwon Ryu}{nc}
\mlsysauthor{Jeongtae Lee}{nc}
\mlsysauthor{Hanbae Seo}{nc}
\mlsysauthor{Hoseung Kim}{nc}
\mlsysauthor{Sunghwan Shim}{nc}
\mlsysauthor{Jaegwang Lee}{nc}
\mlsysauthor{Hyunjoon Jeong}{nc}
\mlsysauthor{Junhee Yoo}{nc}
\mlsysauthor{Sunghyeon Woo}{corr,nc}
\end{mlsysauthorlist}
\mlsysaffiliation{nc}{NAVER Cloud, Seongnam, Republic of Korea}
\mlsyscorrespondingauthor{Sunghyeon Woo}{sunghyeon.woo1@navercorp.com}
\mlsyskeywords{Speculative Decoding, Drafter Training, Compression}

\vskip 0.3in

\begin{abstract}

Recent speculative-decoding drafters such as EAGLE-3 and DFlash learn from the hidden states of the target model. Offline training computes and caches these hidden states once, and therefore needs 2--3$\times$ less GPU time than online training, which recomputes the hidden states at every step. For a standard training corpus, however, the BF16 cache takes 34--55 TB and does not fit on the local disk of an 8-GPU node. The cache can be compressed with methods developed for KV caches, but quantizers that spread bits equally across dimensions substantially shorten the acceptance length of the drafter. We trace this loss to two properties of the cache. First, as in a KV cache, 99\% of the variance of target hidden states lies in 32\% of the principal components, and compression must therefore allocate bits by variance. Second, unlike a KV cache, the cache is training data: the drafter learns without the low-variance components but receives the full hidden states at serving, causing a train--serve mismatch. Reconstruction error is therefore a poor guide for choosing a compression method. Based on these observations, we propose SpecZip, a system for training drafters on a compressed cache. SpecZip compresses the cache with PCA and reverse water-filling (PCA--RWF), which allocates bits by variance without per-token scales and thus keeps more components at the same size, and can additionally project the removed components out of the first layer of the drafter. Because SpecZip decodes only the kept components on the GPU and overlaps the decoding with training, decompression adds almost no time to a training step. A 16$\times$ cache of 2.2--3.4\,TB fits on a single node, and DFlash drafters trained on this cache retain 99--101\% of the BF16 acceptance length across four targets, including a 122B Mixture-of-Experts (MoE) model. Consequently, SpecZip enables offline drafter training from a local disk too small for the uncompressed cache of existing training frameworks, while delivering 1.95--3.25$\times$ higher training throughput than online training.
\end{abstract}

]

\printAffiliationsAndNotice{\mlsysEqualContribution\textsuperscript{\textdagger}Corresponding author }

\section{Introduction}
\label{sec:intro}

\setcounter{figure}{0}
\begin{figure}[!t]
\centering
\includegraphics[width=0.92\columnwidth]{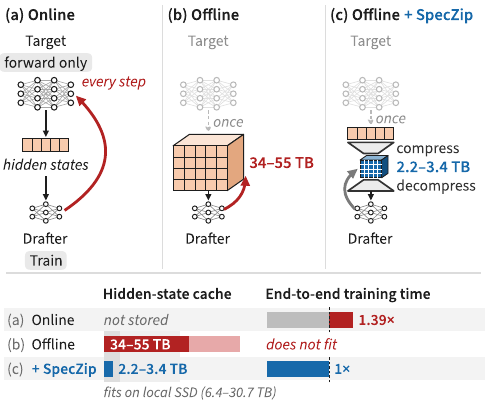}
\caption{Three ways to train a drafter.
(a) Online training recomputes the hidden states at every step.
(b) Offline training computes the hidden states once, but the BF16 cache exceeds the local SSD.
(c) SpecZip stores a 16\texttimes{} compressed cache that fits.}
\label{fig:1}
\end{figure}

Speculative decoding accelerates LLM serving while preserving the output distribution of the target model: a small drafter proposes several tokens, and the target model verifies the proposed tokens in a single forward pass \citep{leviathan2023,chen2023spec}.
Recent drafters such as EAGLE-3 \citep{eagle3}, DFlash \citep{dflash} and DSpark \citep{dspark} condition on the hidden states of the target model and thereby achieve a longer acceptance length $\tau$, the number of tokens accepted per verification step.

\setcounter{table}{0}
\providecommand{\tablefont}{}
\begin{table}[!t]\centering\tablefont\footnotesize\setlength{\tabcolsep}{4pt}
\caption{Acceptance-length retention of Qwen3-4B DFlash drafters at equal bytes.
Under 16\texttimes{} compression, only variance-based bit allocation keeps at least 99\%.}
\label{tab:equal-bytes}
\begin{tabular}{@{}lrrr@{}}
\toprule
Retention (\%) & 16\texttimes{} & 32\texttimes{} & 64\texttimes{} \\
\midrule
\multicolumn{4}{@{}l}{\textit{Equal bits}} \\
\quad TurboQuant & 80.57 & n/a & n/a \\
\quad OPQ & 94.68 & 79.88 & 61.38 \\
\quad PQ & 94.63 & 80.11 & 61.88 \\
\qquad + escape & 94.47 & 79.39 & 62.07 \\
\qquad + PCA basis & 91.02 & 91.52 & 90.36 \\
\cmidrule(l){1-4}
\multicolumn{4}{@{}l}{\textit{Bits by variance}} \\
\quad KVTC & \textbf{99.36} & 97.28 & 94.02 \\
\quad \textbf{PCA--RWF} & \textbf{99.65} & 98.50 & 96.45 \\
\bottomrule
\end{tabular}
\end{table}

Training such a drafter requires the hidden states of several target layers for every training token, and the training data therefore grows with both the number of training tokens and the hidden width of the target. For instance, the DSpark drafters are trained on the DSpark corpus, 1.35M prompts from Open-PerfectBlend that hold more than a billion tokens, and the EAGLE-3 drafters released by the open-source SpecForge framework are trained on 1.4M samples of the same dataset \citep{labonne2024openperfectblend,dspark,specforge}.

Generating these hidden states is computationally costly, because each hidden state requires a full forward pass of the target model. As illustrated in Figure~\ref{fig:1}(a), online training performs this forward pass at every training step \citep{speculators,specforge,torchspec}, and a multi-epoch run therefore recomputes identical hidden states in every epoch. Offline training instead executes the target once and reuses the stored hidden states across epochs, as shown in Figure~\ref{fig:1}(b) \citep{dflash}, and requires 2--3\texttimes{} less GPU time than online training. The stored hidden states, however, impose a substantial storage footprint. For the DSpark corpus, the BF16 cache already reaches 34--55\,TB, exceeding the local SSD of an 8-GPU node. Training must therefore stream the cache from shared storage in every epoch and becomes bound by the shared-storage bandwidth instead of GPU compute. Existing training frameworks nonetheless keep the hidden states uncompressed and treat the storage as an inherent cost of offline training \citep{specforge}.

\setcounter{figure}{1}
\begin{figure}[!t]
\centering
\includegraphics[width=\columnwidth]{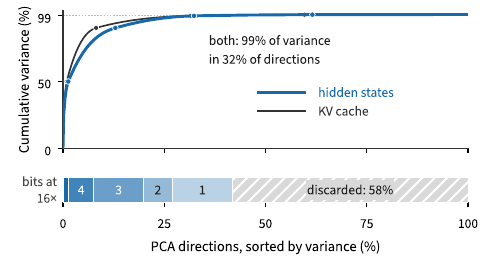}
\caption{Cumulative variance of Qwen3-4B hidden states and KV caches.
Hidden states and KV caches both concentrate 99\% of the variance in 32\% of the principal components, and PCA--RWF under 16\texttimes{} compression keeps 42\% of the components.}
\label{fig:2}
\end{figure}

Tensor compression is well studied in LLM serving, where many methods shrink the KV cache to save GPU memory and bandwidth \citep{kivi,cachegen,kvtc}.
Some methods quantize each value to fewer bits \citep{micikevicius2022fp8,kivi}, and other methods such as KV-cache transform coding (KVTC) first decorrelate the tensor with a rotation or a linear transform \citep{quarot,kvtc,turboquant}.
All of these methods aim to reconstruct the tensor as closely as possible.
Among equally sized hidden-state caches, methods that spread bits equally across dimensions lose a large part of the acceptance length, whereas methods that allocate bits by variance keep nearly the full acceptance length of the BF16 drafter trained on the uncompressed cache, as shown in Table~\ref{tab:equal-bytes}.

We identify two properties of the hidden-state cache that explain this gap. First, target hidden states exhibit the variance structure of a KV cache: hidden states and KV caches concentrate 99\% of the variance in roughly a third of the principal components, as shown in Figure~\ref{fig:2}. A method that spreads bits equally therefore spends most of the byte budget on components that carry little information. Second, the drafter receives the full hidden states at serving, while the drafter is trained on hidden states with the low-variance components removed.
The drafter therefore learns almost nothing about the removed components, and the variation along these components acts as noise at serving, as illustrated in Figure~\ref{fig:3}(a). Under this train--serve mismatch, removing a component costs more acceptance length than quantizing the component coarsely, and a method with lower reconstruction error can still yield a shorter acceptance length. Of the two methods that allocate bits by variance, the method that keeps more components at coarser precision loses less acceptance length at every compression ratio, as shown in Table~\ref{tab:equal-bytes}.

Building on these observations, we present SpecZip, a system for training drafters offline on a compressed hidden-state cache, as summarized in Figure~\ref{fig:1}(c). SpecZip follows two design principles that mirror these observations: \emph{compress like a KV cache} and \emph{see more, ignore the unseen}. During training, SpecZip encodes the cache with PCA and reverse water-filling (PCA--RWF), which allocates bits by variance as KV-cache compression does. Because PCA--RWF stores no per-token scales, the bytes that KVTC spends on scales go to additional components, and PCA--RWF codes 1.3--1.9$\times$ more components than KVTC at the same size, letting the drafter see more of the hidden state. On Qwen3-4B, PCA--RWF loses less acceptance length than KVTC at every compression ratio, even though PCA--RWF reconstructs the hidden states less accurately. At serving, a projection in the first layer of the drafter can further discard the removed components, letting the drafter ignore what training never showed, and the projection applies to any compression method. SpecZip unpacks only the kept components on the GPU and overlaps this unpacking with training on a separate stream, so a training step takes 1.05$\times$ the GPU-only step.

In summary, our contributions are as follows:
\begin{enumerate}
  \item \textbf{Analysis of the train--serve mismatch.}
    To the best of our knowledge, this work is the first to identify and analyze the train--serve mismatch that a compressed hidden-state cache causes in drafter training. We show that the mismatch arises because the drafter is trained without the variation along the removed components but receives this variation as noise at serving.
  \item \textbf{The SpecZip system and its performance.}
  SpecZip compresses the cache 16\texttimes{} without sacrificing the acceptance length. Across targets of various sizes, from a 4B dense model to a 122B Mixture-of-Experts (MoE) model, DFlash drafters trained on the 2.2--3.4\,TB compressed cache retain 99--101\% of the BF16 acceptance length.
  The compressed cache fits on the local disk of a single node, and even on slow shared storage SpecZip trains 1.9\texttimes{} faster than group INT4.
  \item \textbf{Compatibility with existing frameworks.}
  SpecZip integrates into existing serving and training frameworks. The compressed cache is generated directly within vLLM at no additional cost, and the public Speculators trainer runs on the compressed cache with 1.95--3.25\texttimes{} higher training throughput than online training.
\end{enumerate}

\setcounter{figure}{2}
\begin{figure*}[t]
\centering
\includegraphics[width=\textwidth]{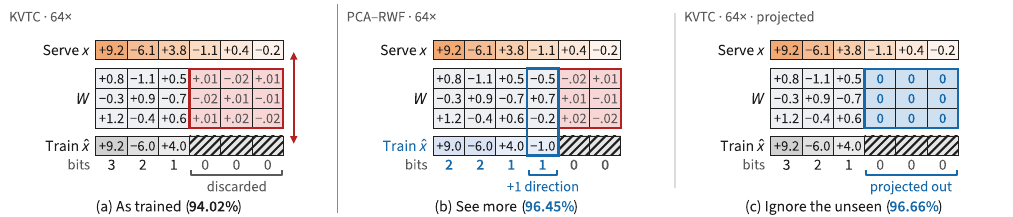}
\caption{See more, ignore the unseen.
Each panel shows what the drafter receives at serving ($x$, top), the first-layer weights of the drafter ($W$, middle) and what the drafter saw in training ($\hat{x}$, bottom), with illustrative values for a Qwen3-4B drafter at 64\texttimes{} compression.
(a) As trained, a drafter trained on a KVTC cache: the last three components receive no bits and look constant in training, and the drafter therefore learns no useful weights for these components (red).
At serving, the same components carry real values, and the untrained weights pass these values on as noise, retaining only 94.02\% of the BF16 acceptance length.
(b) See more, a drafter trained on a PCA--RWF cache: the same bytes cover four components instead of three, leaving only two untrained columns and retaining 96.45\% of the BF16 acceptance length.
(c) Ignore the unseen, the baseline drafter with a projection: setting the untrained columns to zero blocks the noise at serving and retains 96.66\% of the BF16 acceptance length.}
\label{fig:3}
\end{figure*}

\section{Background and Motivation}
\subsection{Speculative Decoding with Target Hidden States}

Recent drafters such as EAGLE-3 \citep{eagle3}, DFlash \citep{dflash} and DSpark \citep{dspark} take as input the hidden states of several target layers, which we call tapped layers.
The drafter concatenates these hidden states and maps the concatenated hidden states to the drafter width with a linear first layer $W$.
Because verification preserves the output distribution of the target \citep{leviathan2023,chen2023spec}, compressing the training data of the drafter can change only the acceptance length $\tau$, not the output.
We therefore report \textbf{retention}, the macro $\tau$ over nine benchmarks of a drafter as a percentage of the $\tau$ of the same drafter trained on BF16 hidden states with the same recipe.

\subsection{Online and Offline Drafter Training}

Online training runs the target in every training step to produce these hidden states \citep{speculators,specforge,torchspec}, and online training therefore repeats the forward pass of the target in every epoch.
Offline training runs the target once, stores the hidden states, and reads the stored hidden states in every epoch \citep{dflash}.
The cost of offline training moves from computation to storage, because the cache grows with the number of training tokens, tapped layers and hidden dimensions.
For the DSpark corpus, the BF16 cache already exceeds the local disk of a training node.

\subsection{LLM Tensor Compression}

We refer to a compression method for the cache as a codec.
Every codec chooses a basis to code in, a way to split bits across coordinates, and whether to store per-token side information \citep{cover2006}.
Scalar formats code each coordinate with a scale per group \citep{micikevicius2022fp8,kivi}, vector quantizers look up codebooks over sub-vectors \citep{jegou2011pq,ge2013opq}, TurboQuant rotates each vector randomly and stores the vector norm \citep{turboquant}, and transform coding encodes the tensor in a decorrelating basis \citep{wallace1992jpeg,kvtc}.

In a decorrelating basis, the classical way to split bits is reverse water-filling \citep{huang1963,cover2006}.
Let $\sigma_i^2$ be the variance of coordinate $i$ and $\theta$ a threshold called the water level.
Reverse water-filling gives coordinate $i$ $\max\!\big(0, \tfrac{1}{2}\log_2(\sigma_i^2/\theta)\big)$ bits, which minimizes the mean-squared error of a Gaussian source for a fixed total number of bits \citep{huang1963,cover2006}.
Coordinates with $\sigma_i^2 \le \theta$ receive no bits and are decompressed to the mean, and we call these coordinates \textbf{discarded}.
KVTC codes KV caches in a PCA basis with bits set by dynamic programming, per-token scales and entropy coding \citep{deflate,kvtc}.

Each of these codecs serves a fixed reader, such as a model or an index that reads the decompressed values, and is judged by reconstruction error or by the accuracy of that reader \citep{kvtc,turboquant}.
A drafter training cache differs: the drafter that reads the cache is trained on decompressed values and then served on exact values.
Reconstruction error therefore need not rank codecs for a training cache.

\section{Compressing the Training Cache}

\textbf{Running setup.} We use a Qwen3-4B target and a DFlash drafter trained with the DeepSpec release recipe for 6 epochs on 100K prompts, a subset of Open-PerfectBlend \citep{labonne2024openperfectblend} that matches its statistics, and evaluate on the nine-benchmark DSpark Table-1 suite (\S{}5.1). Every compressor gets equal bytes per token, from 16\texttimes{} to 64\texttimes{} compression of the 25,600 B that a token takes in BF16. We call a difference \emph{detected} when it lies outside the evaluation noise of that comparison (Appendix~\ref{app:A}) and, on the MoE, also exceeds 1 pp.

\subsection{Compressing Hidden States Like KV Caches}

We observe that the hidden states a drafter reads can be compressed like a KV cache. On the same Qwen3-4B tokens, the hidden states hold 99\% of their variance in 32\% of the directions and the KV cache in about a third (Figure~\ref{fig:2}). Prior work finds the same low-rank structure in activations and keys \citep{ethayarajh2019contextual,slicegpt,espace,loki}, so most directions need few bits or none.

\textbf{Bits should follow variance.} We find that the bit allocation, more than the transform, decides whether the drafter keeps its acceptance. At 16\texttimes{}, only the codecs that allocate bits by variance keep about 99\%, while the codecs that spread bits evenly lose 5--19 pp (Table~\ref{tab:equal-bytes}). With even bits, the large coefficients become coarse, and a better transform does not help: on a PCA basis even bits retain 91.02\%, below plain PQ. On every other target and drafter, the same ordering holds (Table~\ref{tab:retention}), so a training cache needs a codec that allocates bits by variance.

Our default codec, \textbf{PCA--RWF}, is the classical construction of \citet{huang1963}. We apply a mean-removed PCA over the five tapped layers and give each direction 0--8 bits by reverse water-filling \citep{cover2006}. Because the codec stores no per-token side information, every token gets a fixed-length code, and at 16\texttimes{} the codec discards 58\% of the directions (Figure~\ref{fig:2}). Like every codec here, PCA--RWF stores the few extreme tokens at the attention-sink positions exactly (escape) \citep{xiao2024streaming,massiveact}. For comparison, KVTC runs as published on the same transform and escape.

\textbf{Limitations:} Despite keeping about 99\% at 16\texttimes{}, allocating bits by variance still leaves part of the variance in the directions that receive no bits, and retention falls at higher ratios. We need to understand what the drafter loses in these directions.

\setcounter{table}{1}
\providecommand{\tablefont}{}
\begin{table*}[t]\centering\tablefont\footnotesize\setlength{\tabcolsep}{6pt}
\caption{Discarded variance $E_{\perp}$, not NMSE, orders PCA--RWF and KVTC at equal bytes. PCA--RWF codes more directions in every pair and discards less variance, while KVTC has the lower NMSE in 8 of the 9 pairs. PCA--RWF has the higher retention in 8. Retention gap: PCA--RWF minus KVTC. Bold: the better value of each pair and the detected gaps.}
\label{tab:discarded-energy}
\begin{tabular}{@{}llrrrrrrr@{}}
\toprule
& & \multicolumn{2}{c}{Coded directions $\uparrow$} & \multicolumn{2}{c}{$E_{\perp}$ $\downarrow$} & \multicolumn{2}{c}{NMSE $\downarrow$} & Retention gap \\
\cmidrule(lr){3-4}\cmidrule(lr){5-6}\cmidrule(lr){7-8}
Target & Ratio & KVTC & PCA--RWF & KVTC & PCA--RWF & KVTC & PCA--RWF & (pp) \\
\midrule
Qwen3-4B & 16\texttimes{} & 3,872 & \textbf{5,346} & 0.0078 & \textbf{0.0028} & \textbf{0.0130} & 0.0181 & +0.29 \\
 & 32\texttimes{} & 2,336 & \textbf{3,010} & 0.0279 & \textbf{0.0136} & \textbf{0.0529} & 0.0549 & \textbf{+1.21} \\
 & 64\texttimes{} & 1,248 & \textbf{2,245} & 0.0946 & \textbf{0.0321} & \textbf{0.1246} & 0.1672 & \textbf{+2.43} \\
\midrule
Gemma-4-12B & 16\texttimes{} & 6,976 & \textbf{10,944} & 0.0176 & \textbf{0.0037} & 0.0254 & \textbf{0.0239} & \textbf{+0.83} \\
 & 32\texttimes{} & 3,808 & \textbf{6,723} & 0.0420 & \textbf{0.0190} & \textbf{0.0523} & 0.0551 & \textbf{+1.50} \\
 & 64\texttimes{} & 1,936 & \textbf{3,620} & 0.0677 & \textbf{0.0441} & \textbf{0.0768} & 0.0868 & \textbf{+1.70} \\
\midrule
Qwen3.5-122B-A10B & 16\texttimes{} & 4,704 & \textbf{7,424} & 0.0183 & \textbf{0.0068} & \textbf{0.0253} & 0.0356 & \textminus{}0.07 \\
 & 32\texttimes{} & 2,832 & \textbf{3,884} & 0.0457 & \textbf{0.0246} & \textbf{0.0756} & 0.0837 & +0.93 \\
 & 64\texttimes{} & 1,552 & \textbf{2,753} & 0.1228 & \textbf{0.0492} & \textbf{0.1600} & 0.2186 & \textbf{+2.14} \\
\bottomrule
\end{tabular}
\end{table*}

\setcounter{table}{2}
\providecommand{\tablefont}{}
\begin{table}[t]\centering\tablefont\footnotesize\setlength{\tabcolsep}{3pt}
\caption{Seeing more and ignoring the unseen each reduce the train--serve mismatch, and the two are complementary. Parentheses give the change from KVTC as trained, or from PCA--RWF as trained in the last row, for Qwen3-4B DFlash drafters. Bold: detected changes.}
\label{tab:removal}
\begin{tabular*}{\columnwidth}{@{}l@{\extracolsep{\fill}}rrr@{}}
\toprule
Retention (\%) & 16\texttimes{} & 32\texttimes{} & 64\texttimes{} \\
\midrule
\textit{The mismatch:} KVTC, as trained & 99.36 & 97.28 & 94.02 \\
\addlinespace[3pt]
\multicolumn{4}{@{}l}{\textit{See more: discard less}} \\
\quad \textbf{PCA--RWF, as trained} & 99.65 & 98.50 & 96.45 \\[-2.5pt]
 & {\scriptsize (+0.29)} & {\scriptsize (\textbf{+1.21})} & {\scriptsize (\textbf{+2.43})} \\
\addlinespace[3pt]
\multicolumn{4}{@{}l}{\textit{Ignore the unseen: project out (KVTC cache)}} \\
\quad projected training & 99.53 & 97.92 & 96.39 \\[-2.5pt]
 & {\scriptsize (+0.17)} & {\scriptsize (+0.64)} & {\scriptsize (\textbf{+2.37})} \\
\quad zero-init & 99.45 & 97.67 & 96.50 \\[-2.5pt]
 & {\scriptsize (+0.09)} & {\scriptsize (+0.39)} & {\scriptsize (\textbf{+2.48})} \\
\quad serve projected & 99.98 & 98.23 & 96.66 \\[-2.5pt]
 & {\scriptsize (\textbf{+0.62})} & {\scriptsize (\textbf{+0.95})} & {\scriptsize (\textbf{+2.64})} \\
\addlinespace[3pt]
\multicolumn{4}{@{}l}{\textit{Both (SpecZip option)}} \\
\quad \textbf{PCA--RWF + folded projection} & 100.04 & 98.43 & 97.27 \\[-2.5pt]
 & {\scriptsize (+0.39)} & {\scriptsize (\textminus{}0.06)} & {\scriptsize (\textbf{+0.82})} \\
\bottomrule
\end{tabular*}
\end{table}

\subsection{Fixing the Train--Serve Mismatch}

Our second observation is that, unlike a KV cache, the training cache is what the drafter learns from, which creates a train--serve mismatch (Figure~\ref{fig:3}(a)).

\textbf{Analyzing the training error.} We start from the error that a codec leaves in each token. Let S be the subspace of the directions that receive at least one bit, $P_S$ its projector and $\mu$ the PCA mean. The reconstruction error splits into two parts,
\begin{equation}
\begin{aligned} \mathrm{NMSE} &= \frac{\|x - \hat{x}\|^2}{\|x\|^2} = E_{\perp} + E_q, \\ E_{\perp} &= \frac{\|(I - P_S)(x - \mu)\|^2}{\|x\|^2}, \end{aligned}
\end{equation}
where $E_{\perp}$ is the \textbf{discarded variance} and $E_q$ is the quantization noise on the retained directions. On a non-escaped token, $\hat{x}$ equals the mean along every discarded direction, so the discarded directions never change in the cache. During training, the first-layer weights on these directions receive the same gradient as the bias and stay near their initialization. At serving, x does change along the same directions, and the untrained weights turn the change into noise (Figure~\ref{fig:3}(a)). In codec-trained drafters, 75--91\% of that noise comes from the initialization. As a result, the mismatch grows with $E_{\perp}$, while reconstruction error mixes $E_{\perp}$ with $E_q$.

\textbf{To verify the analysis.} We find that the retention of the drafter that a codec trains follows $E_{\perp}$ rather than NMSE (Table~\ref{tab:discarded-energy}). Across three targets, PCA--RWF has the lower $E_{\perp}$ in every pair and the higher retention in 8 of 9, while KVTC has the lower NMSE in 8. On Gemma-4-12B \citep{gemma} at 64\texttimes{}, for example, PCA--RWF leads by 1.70 pp although its NMSE is higher. Within one codec, the same order holds. At 765 B per token, PCA--RWF 32\texttimes{} that codes 4,500 instead of 2,000 directions has the higher NMSE yet retains 98.81\% against 96.12\% (Table~\ref{tab:within-codec}). The drafter thus loses the variance it never saw and should see more directions in training and ignore the unseen ones at serving.

\textbf{See more: discard less.} For the first fix, we spend the same bytes on more directions instead of on more precision, so fewer first-layer weights stay at their initialization (Figure~\ref{fig:3}(a)(b)). PCA--RWF frees these bytes by storing no per-token scales, which KVTC stores for every token. At equal bytes, PCA--RWF codes 1.3--1.9\texttimes{} more directions (Table~\ref{tab:discarded-energy}) and leads KVTC by 2.43 pp at 64\texttimes{} (Table~\ref{tab:removal}). Precision matters far less: removing all quantization noise on the same retained subspace gains only +0.42 pp at 32\texttimes{}. Coding 3,010 instead of 2,000 directions at equal bytes, by contrast, gains 2.38 pp (Table~\ref{tab:within-codec}).

\setcounter{table}{3}
\providecommand{\tablefont}{}
\begin{table}[t]\centering\tablefont\footnotesize\setlength{\tabcolsep}{4pt}
\caption{Within one codec, $E_{\perp}$ orders retention and NMSE does not. At the same 765 B per token, PCA--RWF 32\texttimes{} that codes more directions with fewer bits each discards less variance and keeps at least as much acceptance, while NMSE is lowest at the default $K$ = 3,010. We train Qwen3-4B DFlash drafters and report retention, \% of the BF16 drafter.}
\label{tab:within-codec}
\begin{tabular*}{\columnwidth}{@{\extracolsep{\fill}}rrrrr@{}}
\toprule
Coded $K$ & Bits / dir. & $E_{\perp}$ $\downarrow$ & NMSE $\downarrow$ & Retention (\%) \\
\midrule
2,000 & 3.06 & 0.0444 & 0.0571 & 96.12 \\
3,010 & 2.03 & 0.0136 & 0.0549 & 98.50 \\
4,500 & 1.36 & 0.0051 & 0.0992 & 98.81 \\
\bottomrule
\end{tabular*}
\end{table}

\textbf{Ignore the unseen: project out.} For the second fix, we use the fact that the codec itself defines the discarded directions. We project them out of the drafter's first layer, W $\leftarrow$ W $P_S$, during training or only at serving, so the projection works with any codec (Figure~\ref{fig:3}(c)). The training variants are \emph{projected training} and \emph{zero-init}, and the serving variant is \emph{served projected}. For KVTC at 64\texttimes{}, the projection recovers about 2.5 pp and still helps on PCA--RWF at 64\texttimes{} (Table~\ref{tab:removal}). On a BF16-trained MoE drafter, in contrast, the same projection costs 4.48 pp, so the gain appears only in drafters trained on the cache. In SpecZip the projection is folded into the first layer (\S{}4.2).

\textbf{Advantages.} Neither fix needs a search, since the codec defines S. Seeing more adds no step at training or serving, and ignoring the unseen also applies to drafters that are already trained. SpecZip uses both: it sees more by default, and ignoring the unseen applies to any codec, including KVTC (\S{}4).

\setcounter{figure}{3}
\begin{figure*}[t]
\centering
\includegraphics[width=\textwidth]{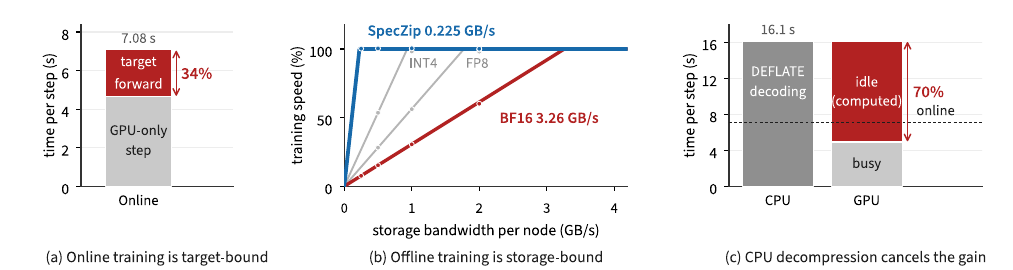}
\caption{Bottleneck analysis of drafter training for Qwen3-4B DFlash on 8\texttimes{} H200. (a) Online training is bound by the target forward. (b) Offline training is bound by storage bandwidth, and SpecZip needs far less of it than BF16, FP8 or INT4. (c) Decompressing a DEFLATE \citep{deflate} cache on the CPU leaves the GPU idle and is slower than online training.}
\label{fig:4}
\end{figure*}

\section{SpecZip: Offline Training from a Compressed Cache}

SpecZip makes a compressed cache as fast to train on as an uncompressed one. Compression alone does not achieve this, because each token must be decompressed before the drafter can use it, and a slow decompression costs more time than the smaller reads save.

\subsection{Why SpecZip Speeds Up Drafter Training}

Each way of training is limited by a different part of the system (Figure~\ref{fig:4}). We profile one training step of a Qwen3-4B DFlash drafter in the DeepSpec trainer \citep{dspark} on 8\texttimes{} H200 and compare it with the GPU-only step, an ideal design that keeps the entire cache in GPU memory.

\textbf{Online training is target-bound.} Online training repeats the same target forward in every epoch, so a ten-epoch run computes the same hidden states ten times while offline training computes them once. As in Figure~\ref{fig:4}(a), the target forward takes a third of the online step, 7.08 s per update against 4.65 s for the GPU-only step. The same holds in the Speculators trainer \citep{speculators} (\S{}5.4), so the gap does not depend on the trainer.

\textbf{Offline training is storage-bound.} Removing the target moves the bottleneck to storage. We cap the bandwidth delivered to the trainer (Figure~\ref{fig:4}(b)), and every format follows the same rule: the step takes the longer of its compute time and the time to read its bytes. A BF16 cache keeps the GPU busy only at 3.26 GB/s per node. The scalar formats of Table~\ref{tab:simpler-formats} do not close this gap, as FP8 still needs 1.77 GB/s and group INT4 0.93 GB/s, and neither fits the local SSD of the smallest node. SpecZip reads 16\texttimes{} fewer bytes than BF16 and needs 0.225 GB/s. Only SpecZip therefore keeps the GPU busy on storage that is too slow for every scalar format.

\textbf{CPU decompression cancels the gain.} Decompressing the cache on the CPU can make offline training slower than online training. We test the usual way to read a compressed cache, which decompresses it in CPU loader workers and copies the decompressed states to the GPU over PCIe. For our port of KVTC with DEFLATE \citep{deflate}, the GPU waits for the CPU during most of the step (Figure~\ref{fig:4}(c)), and the step takes 16.13 s per update against 7.08 s online. The decompression must therefore run on the GPU and stay off the critical path.

\setcounter{figure}{4}
\begin{figure*}[t]
\centering
\includegraphics[width=\textwidth]{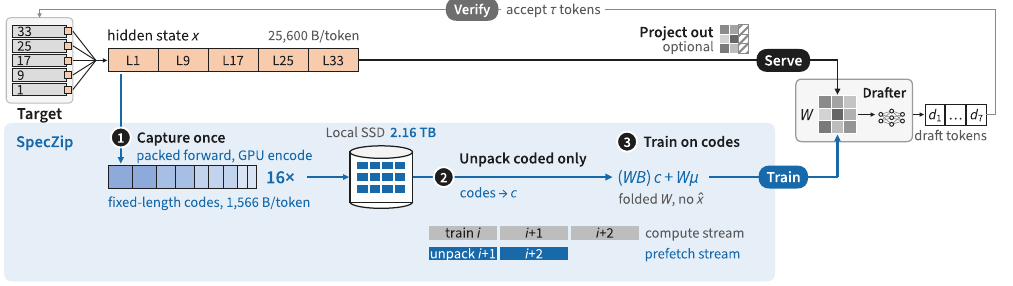}
\caption{Overview of SpecZip. Serving reads the hidden state x unchanged, and training reads the SpecZip cache (blue). (1) The target runs once, and the GPU encodes x into fixed-length codes. (2) A prefetch stream unpacks the next step's coded directions while the current step trains. (3) The folded first layer reads the codes, so training never builds $\hat{x}$, and at serving it can also project out the discarded directions.}
\label{fig:5}
\end{figure*}

\setcounter{table}{4}
\providecommand{\tablefont}{}
\begin{table*}[t]\centering\tablefont\footnotesize\setlength{\tabcolsep}{4.2pt}
\caption{PCA--RWF keeps the BF16 drafter's acceptance at 16\texttimes{} on all four targets, and OPQ at equal bytes (grey) trails in every setting. Each group changes the target or the drafter of Table~\ref{tab:equal-bytes}. Bold: the higher of PCA--RWF and KVTC where the gap is detected. For DSpark and EAGLE-3 the ratio covers only the input views, and they stop at 32\texttimes{}.}
\label{tab:retention}
\begin{tabular}{@{}lrrrrrrrrr@{}}
\toprule
Retention (\%) & \multicolumn{3}{c}{16\texttimes{}} & \multicolumn{3}{c}{32\texttimes{}} & \multicolumn{3}{c}{64\texttimes{}} \\
\cmidrule(lr){2-4}\cmidrule(lr){5-7}\cmidrule(l){8-10}
& \textcolor[HTML]{8A8A8A}{OPQ} & KVTC & PCA--RWF & \textcolor[HTML]{8A8A8A}{OPQ} & KVTC & PCA--RWF & \textcolor[HTML]{8A8A8A}{OPQ} & KVTC & PCA--RWF \\
\midrule
Qwen3-4B DFlash (Table 1) & \textcolor[HTML]{8A8A8A}{94.68} & 99.36 & 99.65 & \textcolor[HTML]{8A8A8A}{79.88} & 97.28 & \textbf{98.50} & \textcolor[HTML]{8A8A8A}{61.38} & 94.02 & \textbf{96.45} \\
\addlinespace[3pt]
\multicolumn{10}{@{}l}{\textit{Other targets (DFlash)}} \\
\quad Llama-3.1-8B & \textcolor[HTML]{8A8A8A}{98.69} & 101.62 & 101.46 & \textcolor[HTML]{8A8A8A}{88.57} & 97.75 & 99.11 & \textcolor[HTML]{8A8A8A}{78.65} & 95.03 & \textbf{99.12} \\
\quad Gemma-4-12B & \textcolor[HTML]{8A8A8A}{96.90} & 98.82 & \textbf{99.65} & \textcolor[HTML]{8A8A8A}{88.18} & 97.98 & \textbf{99.48} & \textcolor[HTML]{8A8A8A}{77.67} & 95.70 & \textbf{97.40} \\
\quad Qwen3.5-122B-A10B & \textcolor[HTML]{8A8A8A}{96.08} & 99.03 & 98.96 & \textcolor[HTML]{8A8A8A}{86.65} & 98.31 & 99.24 & \textcolor[HTML]{8A8A8A}{71.15} & 95.23 & \textbf{97.37} \\
\addlinespace[3pt]
\multicolumn{10}{@{}l}{\textit{Other drafters (Qwen3-4B)}} \\
\quad DSpark & \textcolor[HTML]{8A8A8A}{95.18} & 100.22 & 100.14 & \textcolor[HTML]{8A8A8A}{81.72} & 97.79 & \textbf{98.39} &  &  &  \\
\quad EAGLE-3 & \textcolor[HTML]{8A8A8A}{93.45} & \textbf{98.59} & 96.81 & \textcolor[HTML]{8A8A8A}{80.19} & \textbf{94.64} & 92.44 &  &  &  \\
\bottomrule
\end{tabular}
\end{table*}

\subsection{Training Drafters with SpecZip}

\S{}4.1 leaves two requirements: decompression must run on the GPU and stay off the critical path. To meet both, we design three techniques (Figure~\ref{fig:5}). (1) At capture, the target runs once and the GPU writes x as fixed-length codes. (2) In every training step, the GPU unpacks only the coded directions of the next micro-batch. (3) The drafter's first layer reads these coefficients directly, so no decompressed state $\hat{x}$ is built. We add the techniques one at a time to a cache decompressed in the step (Figure~\ref{fig:8}(b)), and each one shortens the step.

\textbf{Encoding once at capture.} We write the cache once, at a cost close to one epoch of the target forward in online training. We measure the capture in two engines to check that encoding adds no cost (Figure~\ref{fig:8}(a)). In vLLM, encoding on the GPU makes the records 19\texttimes{} smaller at the same cost, and capturing on the trainer's packed forward is another 2.7\texttimes{} cheaper. The whole capture then costs 28.9 GPU-$\mu$s per token, close to the 28.2 GPU-$\mu$s per token of the target forward in an online step. Encoding is cheap because the basis and the bit allocation are fixed before capture, so each token needs one matrix multiplication and a fixed-rate quantization. The 1.35M cache takes 2.16 TB and fits on the smallest node (Table~\ref{tab:simpler-formats}).

\textbf{Unpacking only what is coded.} Unpacking only the coded directions gives the largest saving in the step. Prefetching hides the read behind training yet saves only 0.17 s per update (Figure~\ref{fig:8}(b)). To cut the step further, we skip the full decompression to $\hat{x}$, which saves 1.28 s. Each code has a fixed length, so every direction sits at a known bit offset and unpacking is a small gather on the GPU. The prefetch CUDA stream runs this gather for the next micro-batch in the idle gaps of the current one, so the unpack stays off the critical path of the step.

\textbf{Training directly on codes.} The drafter never needs the decompressed state $\hat{x}$. Because the first layer is linear, W $\hat{x}$ = W $\mu$ + (W B) c, where B is the retained PCA basis and c the coefficients, the first layer can read c directly (Figure~\ref{fig:5}). We fold B and $\mu$ into the first layer and keep W as the parameter. Training on codes changes acceptance by only \textminus{}0.10 pp and runs at the speed of an optimized BF16 reconstruction in the same session (4.87 vs 4.86 s per update). We also fold the projection that ignores the unseen directions (\S{}3.2) into the same GEMM with a bias, and on a MoE drafter this projection changes the time per output token by \textminus{}1.0\% and +0.4\%. The checkpoint stays a standard drafter that existing servers load as is.

\section{Experiments}

\subsection{Settings}

We test four targets: Qwen3-4B \citep{qwen3}, Llama-3.1-8B \citep{llama3}, Gemma-4-12B \citep{gemma} and Qwen3.5-122B-A10B \citep{qwen35}, a MoE with 8 of 256 experts active. DFlash runs on all four targets, and DSpark and EAGLE-3 run on Qwen3-4B. We train on Open-PerfectBlend \citep{labonne2024openperfectblend}, the corpus behind the public EAGLE-3 and DSpark drafters \citep{specforge,dspark}, and take its 1.35M-prompt DeepSpec training split. We follow the DeepSpec release recipe \citep{dspark} at 100K prompts \texttimes{} 6 epochs, and train Qwen3-4B at 1.35M prompts \texttimes{} 10 epochs. We evaluate with DeepSpec \texttt{eval.py} on the nine benchmarks of DSpark Table 1, where the released DFlash checkpoint reproduces the reported $\tau$. We run every experiment on 8\texttimes{} H200. Table~\ref{tab:A1} in Appendix~\ref{app:A} lists every setting.

\setcounter{figure}{5}
\begin{figure*}[t]
\centering
\includegraphics[width=\textwidth]{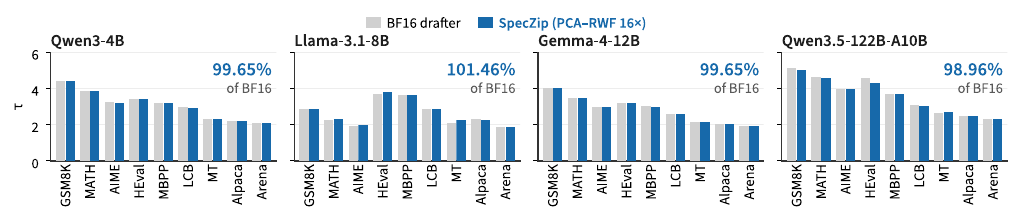}
\caption{Acceptance length per benchmark of BF16 and SpecZip (PCA--RWF 16\texttimes{}) drafters, DFlash on four targets. The blue number in each panel is the retention over the nine benchmarks.}
\label{fig:6}
\end{figure*}

\setcounter{figure}{6}
\begin{figure*}[t]
\centering
\includegraphics[width=\textwidth]{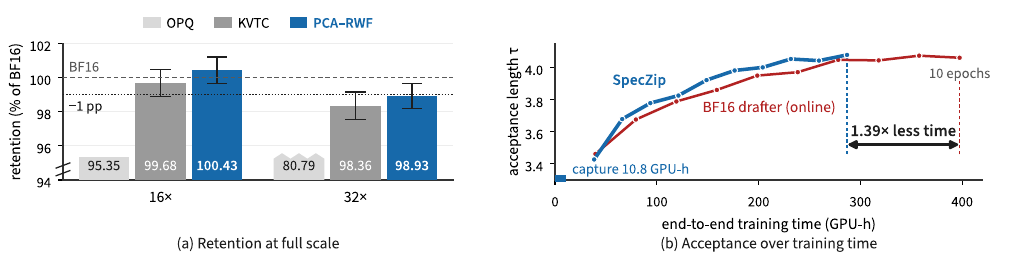}
\caption{At full scale SpecZip keeps the BF16 drafter's acceptance and trains in less end-to-end time. (a) Retention of Qwen3-4B DFlash drafters trained on 1.35M prompts for 10 epochs, against a BF16 drafter trained online with the same recipe. (b) $\tau$ against end-to-end training time. The OPQ bar at 32\texttimes{} is cut at the axis.}
\label{fig:7}
\end{figure*}

\setcounter{figure}{7}
\begin{figure*}[t]
\centering
\includegraphics[width=\textwidth]{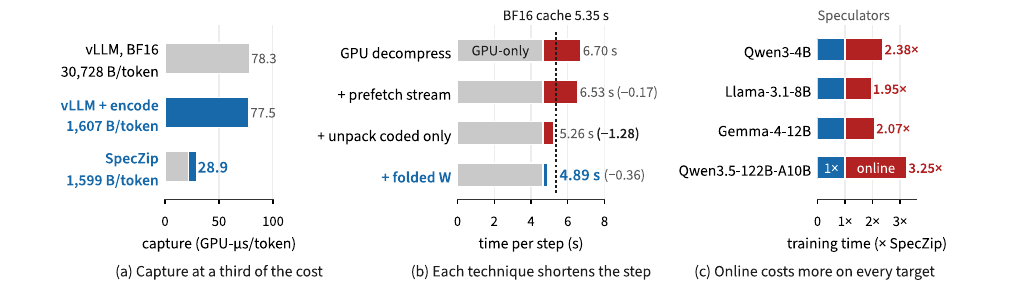}
\caption{Speedup breakdown of SpecZip. (a) Encoding on the GPU and capturing on the trainer's packed forward shorten the capture. (b) Each technique shortens the training step, and SpecZip ends below the BF16-cache step (dashed). (c) In the Speculators trainer \citep{speculators}, online training costs more GPU time per trained token than SpecZip on every target. Blue is SpecZip, red the time a configuration adds, and grey the existing path. Bold: the largest saving.}
\label{fig:8}
\end{figure*}

\subsection{Acceptance Length}

\textbf{Results on dense targets.} SpecZip generalizes across model families and sizes. On Qwen3-4B, Llama-3.1-8B and Gemma-4-12B, which span three families and 4B to 12B parameters, SpecZip retains more than 99\% of the BF16 drafter's acceptance length, and its acceptance length tracks that of the BF16 drafter on all nine benchmarks (Table~\ref{tab:retention}, Figure~\ref{fig:6}, Table~\ref{tab:per-benchmark}). In contrast, OPQ at equal bytes falls behind on every target, by up to 5.0 pp on Qwen3-4B, because it spreads bits equally across dimensions.

\textbf{Results on a MoE target.} For a codec that compresses the target hidden states, a MoE is a harder case. MoE models keep most of their parameters in experts, so their hidden state is narrow for their size \citep{shazeer2017moe,fedus2022switch}, and Qwen3.5-122B-A10B has only 3,072 dimensions, fewer than the 4,096 of Llama-3.1-8B. Even so, SpecZip retains about 99\% of the BF16 drafter's acceptance length at both 16\texttimes{} and 32\texttimes{} (Table~\ref{tab:retention}), because the codec compresses the residual stream that all tokens share, whatever experts they use, and its variance still lies in a few directions.

\textbf{Extreme compression.} The train--serve mismatch is largest at 64\texttimes{}. Every codec discards most directions at this ratio, so more of the variation that the drafter receives at serving was never seen in training (\S{}3.2). SpecZip still retains at least 96\% of the BF16 drafter's acceptance length on every target, and the lead of PCA--RWF over KVTC as trained grows from less than 1\,pp at 16\texttimes{} to 1.7--4.1\,pp at 64\texttimes{}, because PCA--RWF codes more directions at equal bytes (Table~\ref{tab:retention}). KVTC has the lower NMSE in 8 of the 9 pairs of Table~\ref{tab:discarded-energy}, so the lead follows the coded directions, which NMSE does not measure. Ignoring the unseen helps KVTC at this ratio as well: with the folded projection of SpecZip, KVTC gains 1.9--4.0\,pp at 64\texttimes{} and retains at least 96.6\% on every target (Table~\ref{tab:removal} for Qwen3-4B).

\textbf{Other drafters.} SpecZip carries over to drafters that read the target hidden states differently, because the codec changes only those hidden states. DSpark, which adds a Markov head and a different loss to DFlash, retains about 100\% of the BF16 drafter's acceptance length at 16\texttimes{} (Table~\ref{tab:retention}). EAGLE-3 is a harder case. EAGLE-3 feeds its own hidden state back at each drafting step, so an error in its input accumulates over the steps \citep{eagle,eagle3}. On EAGLE-3, KVTC as trained leads PCA--RWF by 1.8\,pp at 16\texttimes{} and 2.2\,pp at 32\texttimes{} (Table~\ref{tab:retention}). When EAGLE-3 is trained on the 3,010 directions that PCA--RWF codes at 32\texttimes{} but without quantization noise, it retains 100.65\%, against 92.44\% with PCA--RWF. The discarded directions thus barely affect EAGLE-3, and the loss comes from quantization noise. KVTC codes fewer directions with more bits each and has the lower NMSE (Table~\ref{tab:discarded-energy}), so EAGLE-3 is better trained on a KVTC cache, which SpecZip also stores.

\setcounter{table}{5}
\providecommand{\tablefont}{}
\begin{table}[t]\centering\tablefont\footnotesize\setlength{\tabcolsep}{4pt}
\caption{Only SpecZip fits the local SSD of every 8-GPU node, and it retains about as much acceptance as the scalar formats. We size the Qwen3-4B cache for 1.35M prompts against the 6.4 TB SSD of the smallest node \citep{googlecloudgpus,amazonp5}, and measure retention at 100K prompts. Bold: SpecZip.}
\label{tab:simpler-formats}
\begin{tabular}{@{}lrrcr@{}}
\toprule
& & Cache & Fits & Retention \\
Format & Ratio & (TB) & SSD & (\%) \\
\midrule
BF16 & 1\texttimes{} & 34.44 & no & 100.00 \\
FP8 & 2.0\texttimes{} & 17.25 & no & 100.29 \\
group INT4 & 3.9\texttimes{} & 8.88 & no & 99.75 \\
\textbf{SpecZip (PCA--RWF 16\texttimes{})} & \textbf{16\texttimes{}} & \textbf{2.16} & \textbf{yes} & \textbf{99.65} \\
\bottomrule
\end{tabular}
\end{table}

\subsection{Full-Scale Training}

\textbf{Results.} The 100K runs train on a small corpus for few updates, so their result might not hold at full scale. The full-scale run therefore trains Qwen3-4B on all 1.35M prompts of the DeepSpec split for 10 epochs, with 22\texttimes{} more updates, at the scale of public drafters (\S{}1). At this scale SpecZip retains 100.4\% of the online BF16 drafter's acceptance length at 16\texttimes{}, above its 99.65\% at 100K, and PCA--RWF still leads KVTC as trained (Figure~\ref{fig:7}(a)). The 100K result thus holds at full scale, and the loss from compression shrinks as training grows.

\textbf{Acceptance over training time.} SpecZip shortens end-to-end training time even on a small target. Qwen3-4B is small, so its target forward is only about a third of the online step and drafter training dominates. Even so, with the one-time capture included and $\tau$ matched throughout, SpecZip overtakes online training within the first epoch and needs 1.39\texttimes{} less end-to-end training time over 10 epochs (Figure~\ref{fig:7}(b)). The saving comes from removing the target forward, so it grows with the share of the step that the forward takes.

\setcounter{table}{6}
\providecommand{\tablefont}{}
\begin{table}[t]\centering\tablefont\footnotesize\setlength{\tabcolsep}{4pt}
\caption{The full-scale SpecZip drafters of Figure~\ref{fig:7} serve as fast as the BF16 drafter in vLLM. Each cell is the speedup over autoregressive (AR) decoding, averaged over the nine benchmarks (Table~\ref{tab:serving-speed}), at temperature T and concurrency c.}
\label{tab:serving}
\begin{tabular}{@{}llrrrr@{}}
\toprule
 & & AR & BF16 & \multicolumn{2}{c@{}}{SpecZip (PCA--RWF)} \\
\cmidrule(l){5-6}
Temp. & c & (tok/s) & drafter & 16\texttimes{} & 32\texttimes{} \\
\midrule
1 & 1 & 281 & 3.11\texttimes{} & 3.15\texttimes{} & 3.17\texttimes{} \\
 & 8 & 2,026 & 2.91\texttimes{} & 2.93\texttimes{} & 2.89\texttimes{} \\
 & 32 & 6,331 & 2.28\texttimes{} & 2.27\texttimes{} & 2.24\texttimes{} \\
\midrule
0 & 1 & 289 & 3.54\texttimes{} & 3.51\texttimes{} & 3.48\texttimes{} \\
 & 8 & 2,084 & 3.29\texttimes{} & 3.31\texttimes{} & 3.25\texttimes{} \\
 & 32 & 6,483 & 2.64\texttimes{} & 2.60\texttimes{} & 2.62\texttimes{} \\
\bottomrule
\end{tabular}
\end{table}

\subsection{Training Speed and Cost}

\textbf{Results.} SpecZip makes offline training possible on nodes too small for the BF16 cache, and decompression adds almost no time to a training step. The 16\texttimes{} cache takes 2.2--3.4\,TB on the four targets and fits on the local SSD of even the smallest 8-GPU node, where every scalar format overflows (Table~\ref{tab:simpler-formats}). In the DeepSpec trainer, the SpecZip step is 1.05\texttimes{} the GPU-only step, because the GPU unpacks only the coded directions, off the critical path (Figure~\ref{fig:8}(b)). In the Speculators trainer, SpecZip delivers 1.95--3.25\texttimes{} higher training throughput than online training on four targets, measured as GPU time per trained token, and stays within 2\% of an offline BF16 cache (Figure~\ref{fig:8}(c)). Training from a compressed cache is thus almost as fast as training from one held in GPU memory.

\textbf{Comparisons against other systems.} Group INT4 is the simplest alternative to SpecZip, and with plentiful bandwidth it matches the SpecZip step (Figure~\ref{fig:4}(b)). Group INT4 reads about four times the bytes, however, so its 1.35M cache does not fit the smallest node, and when the cache streams from shared storage at 0.5\,GB/s per node, its step takes 1.9\texttimes{} as long (Figure~\ref{fig:4}(b)). SpecZip keeps both its cache and its step small on slow storage.

\textbf{Decoding speed.} Acceptance length does not include the cost of drafting and verification, so the full-scale drafters are also served in vLLM at temperatures 0 and 1 and concurrency 1, 8 and 32 (Table~\ref{tab:serving}). The SpecZip drafter at 16\texttimes{} speeds up decoding by 2.27--3.51\texttimes{} over autoregressive decoding, within 1.5\% of the BF16 drafter in every setting and ahead of it in three of the six. At 32\texttimes{} the drafter retains about 99\% of the BF16 drafter's acceptance length, and its decoding speed still stays within 2\% of the BF16 drafter and ahead of it in one setting (Table~\ref{tab:serving-speed}), so a small loss in acceptance length barely shows in decoding speed. Serving needs no codec and no decompression kernel, because the served drafter reads the target's hidden states x directly, so the smaller training cache comes with the same serving speed.

\section{Related Work}

\textbf{Drafters and trainers.} Recent drafters read the hidden states of several target layers \citep{eagle,eagle3,dflash,dspark}, and HASS aligns their training and decoding contexts \citep{hass25}. Existing trainers either recompute these hidden states in every step or compute them once and store them \citep{speculators,specforge,torchspec}. TIDE \citep{tide} instead reuses the hidden states produced during serving. SpecZip compresses the stored hidden states.

\textbf{Compressing KV caches and activations.} KV-cache compressors quantize, rotate or transform-code a cache \citep{kivi,kvquant,cachegen,kvtc,turboquant}. Other work projects activations and keys onto a few principal components, because most of their variance lies in these directions \citep{espace,slicegpt,loki,eigenattention,palu,xkv}, and activation-compressed training stores transient activations in a few bits \citep{actnn,gact}. SpecZip borrows these tools, but these methods assume a reader that does not change. A drafter, in contrast, is trained on decompressed values and served on exact ones. The closest prior caches for training store teacher signals for distillation, such as BERT hidden states \citep{liu2021bert} or sparse logits \citep{sparselogit}, as targets that a student imitates. A student never reads these targets at serving. A drafter, however, reads exact hidden states at serving, so a cache of its inputs creates the train--serve mismatch.

\textbf{Transform coding and implicit bias.} Our codec reuses the classical transform code of \citet{huang1963} with reverse water-filling \citep{cover2006}, and product quantization \citep{jegou2011pq,ge2013opq} serves as our equal-bit baseline. Our analysis of the mismatch also builds on known theory. Gradient descent leaves the part of a linear layer outside the span of its training inputs at initialization \citep{gunasekar2018}. The discarded directions of a codec hold only their mean in the cache, so the first-layer weights on these directions receive only the gradient of the bias and stay near their initial values.

\textbf{Input pipelines.} Input-pipeline work removes data stalls \citep{coordl,ffcv}, and DALI \citep{dali} and nvCOMP \citep{nvcomp} decode images and general data on the GPU. SpecZip also decompresses on the GPU and hides this work behind training. The closest prior system is PCR \citep{pcr}, because it also trains on lossy compressed data. However, PCR decodes each image back to pixels, whereas SpecZip never rebuilds the hidden state and feeds the codes directly to the drafter's first layer.

\section{Conclusion}
In this work, we propose SpecZip, a simple yet effective system for training speculative-decoding drafters on a compressed cache of target hidden states. Based on the observation that target hidden states, like KV caches, hold 99\% of their variance in 32\% of the directions, SpecZip gives each direction bits by its variance. Unlike a KV cache, the cache trains the drafter that reads it, so SpecZip codes more directions than KVTC at the same size and can project the directions without bits out of the drafter's first layer. As a result, SpecZip keeps 99--101\% of the BF16 drafter's acceptance length on four targets, including a 122B Mixture-of-Experts model, and 100\% at full scale. It also applies to other drafter designs such as DSpark and EAGLE-3. Its cache of 2.2--3.4\,TB fits on the local disk of a single node. SpecZip further keeps the speedup of offline training on this compressed cache, because it unpacks only the coded directions on the GPU and overlaps the unpacking with training. Compared with online training, a full Qwen3-4B run needs 1.39\texttimes{} less end-to-end training time including the one-time capture, and the Speculators trainer delivers 1.95--3.25\texttimes{} higher training throughput on four targets.

\bibliographystyle{mlsys2025}
\bibliography{references}

\section*{AI Usage Disclosure}

We used two AI agents throughout this work: Claude (Anthropic), through Claude Code with Claude Opus models, and OpenAI GPT models, through the Codex CLI. We describe their use for each part of the work.

\textbf{Ideation and literature search.} We discussed research directions and candidate explanations with both agents and used them to search for and summarize related work. The authors chose the research questions, the hypotheses to test and the claims of the paper. The authors checked every cited work against its original source.

\textbf{Experiment design and code.} The authors designed the experiments, including the arms, the comparisons, the training budgets and the evaluation protocol. The agents wrote most of the code under the authors' instructions: the codecs, cache capture, training launchers, evaluation harnesses and analysis scripts. They also launched and monitored jobs on our GPU cluster. The authors reviewed the code and checked the results.

\textbf{Data and numbers.} No AI tool generated the training data, other than the target models under study, which generate the training responses. Every number in the paper comes from our experiment logs and is read from a single registry of results. The agents generated the figures and tables from this registry with scripts, and checked that the text, tables and figures match it. No AI tool produced or edited a measured value.

\textbf{Writing.} The authors set the outline, the claims and the abstract. The agents drafted and rewrote paragraphs from the authors' outline and instructions, edited the language, laid out the LaTeX, and acted as internal reviewers of drafts. The authors revised and approved all text.

The authors take full responsibility for the content of this paper, including any content produced with AI assistance.

\clearpage
\onecolumn
\appendix
\renewenvironment{table*}[1][]{\begin{table}[H]}{\end{table}}
\renewenvironment{figure*}[1][]{\begin{figure}[H]}{\end{figure}}

\section{Reproducibility}\label{app:A}
\renewcommand{\thetable}{A\arabic{table}}
\renewcommand{\thefigure}{A\arabic{figure}}

\setcounter{table}{0}
\begin{table*}[t]\centering\footnotesize
\caption{Reproducibility settings. The first rows are shared by every arm, and the last rows give each compressor its own settings.}\label{tab:A1}
\begin{tabularx}{\textwidth}{@{}lX@{}}
\toprule
item & setting \\
\midrule
Training corpus & Open-PerfectBlend \citep{labonne2024openperfectblend}, 1.35M-prompt DeepSpec training split (95\% of the 1.42M rows) with responses regenerated by each target (1,345,413,651 tokens for Qwen3-4B \citep{qwen3}); 100K runs use a fixed 100K-prompt subset that matches the statistics of this split (96,065 matched for Gemma-4-12B \citep{gemma}) \\
Tapped layers & Five layers at round(linspace(1, L \textminus{} 3, 5)): Qwen3-4B \citep{qwen3} 1, 9, 17, 25, 33; Qwen3.5-122B-A10B \citep{qwen35} 1, 12, 23, 34, 45 \\
Recipe & DeepSpec \citep{dspark} release: AdamW \citep{loshchilov2019decoupled}, lr 6e-4, warmup 0.04, global batch 512, weight decay 0, block 7; 100K \texttimes{} 6 epochs = 1,170 updates; 1.35M \texttimes{} 10 = 26,160 \\
Codec fit & PCA \citep{huang1963} mean and basis over the five concatenated views, 131K tokens of training documents, fp64 moments \\
Escape rule & Norm over 8 attention-sink coordinates \citep{xiao2024streaming,massiveact} above 1,188.8 (Qwen3-4B \citep{qwen3}) $\rightarrow$ BF16 row; 0.12\% of tokens, 27.9 B per token \\
PCA--RWF & Reverse water-filling \citep{cover2006} on PCA \citep{huang1963} variances, 0--8 bits per direction, uniform quantizer on [\textminus{}3$\sigma$, 3$\sigma$]; 1,566 B of codes per non-escaped token at 16\texttimes{} \\
KVTC & Paper quantizer \citep{kvtc}: DP allocation over coordinate blocks minimizing calibration MSE, per-token block min--max int2 / int4 or FP8, DEFLATE \citep{deflate} at the byte budget; our escape rule added \\
PQ \citep{jegou2011pq} / OPQ \citep{ge2013opq} & k = 256, u8 codes, 8-dimensional sub-vectors at 16\texttimes{} (16-dimensional at 32\texttimes{}); OPQ \citep{ge2013opq} learns a rotation; no escape unless stated \\
Uniform bits & Our codec's mean, PCA \citep{huang1963} basis and escape rule; the same bit budget (8 \texttimes{} 1,566 bits) spread equally over the components in eigenvalue order, which at 16\texttimes{} gives 1 bit to each of the top 12,528 of 12,800 directions and discards the last 272; scale chosen among 1.6$\sigma$ / 3$\sigma$ / 4$\sigma$ / training range by training-sample MSE \\
TurboQuant & Algorithm 1 \citep{turboquant} (normalize, random rotation, scalar codebook, rescale; per-vector norm stored); 15.91\texttimes{} at "16\texttimes{}" (not exactly byte-matched) \\
Evaluation & DeepSpec \citep{dspark,deepseekdeepspec} \texttt{eval.py}, DSpark Table-1 suite (nine benchmarks, 3,030 prompts; Table~\ref{tab:per-benchmark}), temperature 1.0, block 7; paired bootstrap within benchmark \citep{koehn2004bootstrap}, 10,000 draws \\
Hardware & Every experiment on 8\texttimes{} H200: training, acceptance (MoE in vLLM 0.31 \citep{vllm,vllm2026release}, TP 8) and every timing (Figures~\ref{fig:1}, \ref{fig:4}, \ref{fig:7} and~\ref{fig:8}, \S{}4.2, \S{}5.4) \\
Byte accounting & Ratios exclude the 5 B per token of token ids and loss mask; the codec (655.6 MB) is stored once \\
\bottomrule
\end{tabularx}

\end{table*}

\setcounter{table}{1}
\providecommand{\tablefont}{}
\begin{table*}[t]\centering\tablefont\footnotesize\setlength{\tabcolsep}{4pt}
\caption{Acceptance length per benchmark for Figure~\ref{fig:6}. Each row is one DFlash drafter trained on 100K prompts, with $\tau$ on the nine benchmarks, the macro average and the retention. The benchmarks are GSM8K \citep{cobbe2021verifiers}, MATH-500 \citep{lightman2024verify}, AIME 2025 \citep{matharena2025aime}, HumanEval \citep{chen2021code}, MBPP \citep{austin2021program}, LiveCodeBench \citep{jain2025livecodebench}, MT-Bench \citep{zheng2023judging}, Alpaca \citep{taori2023alpaca} and Arena-Hard-v2 \citep{li2025arenahard,lmarena2025arenahardv2}.}
\label{tab:per-benchmark}
\begin{tabular*}{\textwidth}{@{\extracolsep{\fill}}lrrrrrrrrrrr@{}}
\toprule
Drafter & \makebox[2.6em][r]{GSM8K} & \makebox[2.6em][r]{MATH} & \makebox[2.6em][r]{AIME} & \makebox[2.6em][r]{HEval} & \makebox[2.6em][r]{MBPP} & \makebox[2.6em][r]{LCB} & \makebox[2.6em][r]{MT} & \makebox[2.6em][r]{Alpaca} & \makebox[2.6em][r]{Arena} & \makebox[2.6em][r]{Mean} & Retention (\%) \\
\midrule
\multicolumn{12}{@{}l}{\emph{Qwen3-4B}} \\
\quad BF16 drafter & 4.39 & 3.87 & 3.28 & 3.40 & 3.21 & 2.98 & 2.31 & 2.20 & 2.11 & 3.08 &  \\
\quad SpecZip (PCA--RWF 16\texttimes{}) & 4.43 & 3.83 & 3.21 & 3.39 & 3.20 & 2.95 & 2.33 & 2.19 & 2.12 & 3.07 & 99.65 \\
\midrule
\multicolumn{12}{@{}l}{\emph{Llama-3.1-8B}} \\
\quad BF16 drafter & 2.85 & 2.26 & 1.92 & 3.68 & 3.64 & 2.87 & 2.12 & 2.30 & 1.88 & 2.61 &  \\
\quad SpecZip (PCA--RWF 16\texttimes{}) & 2.88 & 2.31 & 2.01 & 3.80 & 3.65 & 2.84 & 2.25 & 2.27 & 1.85 & 2.65 & 101.46 \\
\midrule
\multicolumn{12}{@{}l}{\emph{Gemma-4-12B}} \\
\quad BF16 drafter & 4.03 & 3.49 & 2.98 & 3.19 & 3.00 & 2.61 & 2.15 & 2.06 & 1.90 & 2.82 &  \\
\quad SpecZip (PCA--RWF 16\texttimes{}) & 4.00 & 3.48 & 2.97 & 3.18 & 2.98 & 2.59 & 2.17 & 2.06 & 1.90 & 2.81 & 99.65 \\
\midrule
\multicolumn{12}{@{}l}{\emph{Qwen3.5-122B-A10B}} \\
\quad BF16 drafter & 5.11 & 4.61 & 3.98 & 4.57 & 3.67 & 3.08 & 2.66 & 2.46 & 2.33 & 3.61 &  \\
\quad SpecZip (PCA--RWF 16\texttimes{}) & 5.04 & 4.60 & 3.98 & 4.32 & 3.68 & 3.05 & 2.68 & 2.47 & 2.32 & 3.57 & 98.96 \\
\bottomrule
\end{tabular*}
\end{table*}

\providecommand{\tablefont}{}
\begin{table*}[t]\centering\tablefont\footnotesize\setlength{\tabcolsep}{4pt}
\caption{Serving speed per benchmark for Table~\ref{tab:serving} (Qwen3-4B DFlash drafters of Figure~\ref{fig:7}, vLLM, one H200 per server, output up to 1,024 tokens, mean of three rounds). In each group the AR row gives output tokens per second without a drafter, and the drafter rows give the speedup over it. The per-benchmark runs placed eight servers on one node, and All is one run on the nine benchmarks together, one server per node. One Arena-Hard prompt exceeds the 8,192-token context at concurrency 8 and 32 and fails in every run.}
\label{tab:serving-speed}
\begin{tabular*}{\textwidth}{@{\extracolsep{\fill}}lrrrrrrrrrrr@{}}
\toprule
Drafter & \makebox[2.6em][r]{GSM8K} & \makebox[2.6em][r]{MATH} & \makebox[2.6em][r]{AIME} & \makebox[2.6em][r]{HEval} & \makebox[2.6em][r]{MBPP} & \makebox[2.6em][r]{LCB} & \makebox[2.6em][r]{MT} & \makebox[2.6em][r]{Alpaca} & \makebox[2.6em][r]{Arena} & \makebox[2.6em][r]{Mean} & All \\
\midrule
\multicolumn{12}{@{}l}{\emph{Temperature 1, concurrency 1}} \\
\quad AR (tok/s) & 281 & 281 & 280 & 281 & 283 & 277 & 281 & 281 & 281 & 281 & 273 \\
\quad BF16 drafter & 4.08 & 3.56 & 3.10 & 3.56 & 3.53 & 3.42 & 2.01 & 2.38 & 2.36 & 3.11 & 3.03 \\
\quad SpecZip (PCA--RWF 16\texttimes{}) & 4.10 & 3.65 & 3.12 & 3.65 & 3.47 & 3.45 & 2.03 & 2.43 & 2.41 & 3.15 & 3.03 \\
\quad SpecZip (PCA--RWF 32\texttimes{}) & 4.16 & 3.66 & 3.21 & 3.61 & 3.52 & 3.30 & 2.07 & 2.47 & 2.50 & 3.17 & 2.96 \\
\midrule
\multicolumn{12}{@{}l}{\emph{Temperature 1, concurrency 8}} \\
\quad AR (tok/s) & 2,127 & 2,089 & 1,906 & 2,033 & 2,097 & 1,878 & 1,976 & 2,105 & 2,021 & 2,026 & 2,037 \\
\quad BF16 drafter & 3.78 & 3.59 & 2.72 & 3.37 & 3.07 & 2.95 & 2.37 & 2.22 & 2.10 & 2.91 & 2.80 \\
\quad SpecZip (PCA--RWF 16\texttimes{}) & 3.80 & 3.61 & 2.85 & 3.34 & 3.09 & 2.98 & 2.39 & 2.23 & 2.13 & 2.93 & 2.76 \\
\quad SpecZip (PCA--RWF 32\texttimes{}) & 3.74 & 3.59 & 2.83 & 3.26 & 3.04 & 2.88 & 2.34 & 2.23 & 2.09 & 2.89 & 2.75 \\
\midrule
\multicolumn{12}{@{}l}{\emph{Temperature 1, concurrency 32}} \\
\quad AR (tok/s) & 7,064 & 6,896 & 6,312 & 6,236 & 6,234 & 5,831 & 5,286 & 6,817 & 6,306 & 6,331 & 6,618 \\
\quad BF16 drafter & 2.81 & 2.84 & 2.00 & 2.54 & 2.55 & 2.48 & 1.86 & 1.76 & 1.70 & 2.28 & 2.08 \\
\quad SpecZip (PCA--RWF 16\texttimes{}) & 2.79 & 2.85 & 1.94 & 2.58 & 2.46 & 2.49 & 1.86 & 1.74 & 1.71 & 2.27 & 2.06 \\
\quad SpecZip (PCA--RWF 32\texttimes{}) & 2.87 & 2.80 & 1.85 & 2.47 & 2.42 & 2.46 & 1.91 & 1.73 & 1.66 & 2.24 & 2.04 \\
\midrule
\multicolumn{12}{@{}l}{\emph{Temperature 0, concurrency 1}} \\
\quad AR (tok/s) & 290 & 289 & 289 & 289 & 291 & 284 & 289 & 289 & 288 & 289 & 282 \\
\quad BF16 drafter & 4.50 & 4.28 & 3.96 & 3.95 & 3.78 & 3.69 & 2.24 & 2.56 & 2.87 & 3.54 & 3.42 \\
\quad SpecZip (PCA--RWF 16\texttimes{}) & 4.42 & 4.29 & 3.93 & 3.90 & 3.81 & 3.65 & 2.23 & 2.54 & 2.80 & 3.51 & 3.42 \\
\quad SpecZip (PCA--RWF 32\texttimes{}) & 4.44 & 4.23 & 3.90 & 3.92 & 3.70 & 3.62 & 2.23 & 2.54 & 2.73 & 3.48 & 3.37 \\
\midrule
\multicolumn{12}{@{}l}{\emph{Temperature 0, concurrency 8}} \\
\quad AR (tok/s) & 2,188 & 2,159 & 1,963 & 2,105 & 2,099 & 1,943 & 2,050 & 2,166 & 2,084 & 2,084 & 2,103 \\
\quad BF16 drafter & 4.14 & 4.21 & 3.57 & 3.68 & 3.44 & 3.20 & 2.53 & 2.39 & 2.41 & 3.29 & 3.13 \\
\quad SpecZip (PCA--RWF 16\texttimes{}) & 4.17 & 4.22 & 3.68 & 3.67 & 3.50 & 3.20 & 2.53 & 2.39 & 2.41 & 3.31 & 3.11 \\
\quad SpecZip (PCA--RWF 32\texttimes{}) & 4.18 & 4.17 & 3.44 & 3.58 & 3.44 & 3.19 & 2.51 & 2.35 & 2.35 & 3.25 & 3.07 \\
\midrule
\multicolumn{12}{@{}l}{\emph{Temperature 0, concurrency 32}} \\
\quad AR (tok/s) & 7,475 & 7,073 & 6,336 & 6,118 & 6,448 & 6,065 & 5,437 & 6,951 & 6,446 & 6,483 & 6,803 \\
\quad BF16 drafter & 3.15 & 3.35 & 2.61 & 3.05 & 2.79 & 2.85 & 2.07 & 1.93 & 1.99 & 2.64 & 2.34 \\
\quad SpecZip (PCA--RWF 16\texttimes{}) & 3.17 & 3.34 & 2.48 & 3.04 & 2.66 & 2.80 & 2.08 & 1.92 & 1.96 & 2.60 & 2.33 \\
\quad SpecZip (PCA--RWF 32\texttimes{}) & 3.21 & 3.32 & 2.56 & 2.96 & 2.83 & 2.79 & 2.09 & 1.90 & 1.94 & 2.62 & 2.30 \\
\bottomrule
\end{tabular*}
\end{table*}

\end{document}